\documentclass[journal]{IEEEtran}
\usepackage{cite}
\usepackage{amsmath,amssymb,amsfonts}
\usepackage{algorithmic}
\usepackage{graphicx}
\usepackage{textcomp}
\usepackage{xcolor}
\usepackage{amsthm}
\usepackage{subfigure}
\usepackage{graphicx}
\usepackage{color}
\usepackage{setspace}
\usepackage{enumerate}

\newtheorem{myDef}{Definition}
\newtheorem{myRemarks}{Remarks}
\newtheorem{myDemo}{Example}

\begin{document}

\title{A Comprehensive Survey of Incentive Mechanism for Federated Learning}

\author{
\IEEEauthorblockN{Rongfei Zeng\IEEEauthorrefmark{1}\IEEEauthorrefmark{4},
Chao Zeng\IEEEauthorrefmark{1},
Xingwei Wang\IEEEauthorrefmark{2},
Bo Li\IEEEauthorrefmark{3},
and Xiaowen Chu\IEEEauthorrefmark{4}}\\
\IEEEauthorblockA{\IEEEauthorrefmark{1} Software College, Northeastern University, Shenyang, China}\\
\IEEEauthorblockA{\IEEEauthorrefmark{2} Department of Computer Science, Northeastern University, Shenyang, China}\\
\IEEEauthorblockA{\IEEEauthorrefmark{3} Department of Computer Science and Engineering, HKUST, Hong Kong, China}\\
\IEEEauthorblockA{\IEEEauthorrefmark{4} Department of Computer Science, Hong Kong Baptist University, Hong Kong, China}\\
\IEEEauthorblockA{Email: zengrf@swc.neu.edu.cn, zengc@gmail.com, wangxw@neu.edu.cn, bli@ust.hk, chxw@comp.hkbu.edu.hk}
}

\maketitle

\begin{abstract}
Federated learning utilizes various resources provided by participants to collaboratively train a global model, which potentially address the data privacy issue of machine learning. In such promising paradigm, the performance will be deteriorated without sufficient training data and other resources in the learning process. Thus, it is quite crucial to inspire more participants to contribute their valuable resources with some payments for federated learning. In this paper, we present a comprehensive survey of incentive schemes for federate learning. Specifically, we identify the incentive problem in federated learning and then provide a taxonomy for various schemes. Subsequently, we summarize the existing incentive mechanisms in terms of the main techniques, such as Stackelberg game, auction, contract theory, Shapley value, reinforcement learning, blockchain. By reviewing and comparing some impressive results, we figure out three directions for the future study. 

\end{abstract}

\section{Introduction}
As a promising distributed deep learning paradigm, Federated Learning (FL) is proposed to collaboratively train a global machine learning model with plenty of participants whose data privacy is considered as top priority \cite{Google17:McMahan}. In FL, each participant trains a local model with its private data samples and then submits model parameters to the remote cloud instead of raw training data. After collecting sufficient parameters of local models, a global model is aggregated and then distributed to participants for next round of local training. The process iterates until the global model satisfies the predefined accuracy requirement. From the training process, we can easily learn that FL drastically improves the data privacy of participants, without the upload of raw data.  

The salient feature of FL enables its widespread applications in both cross-device and cross-silo settings. In cross-device FL, more clients are fascinated to contribute their resources to improve their user experience. For example, Google applies FL to its products Gboard to improve the performance \cite{Others18:Hard}. Similarly, Apple employs FL to QuickType and ``Hey Siri" of iOS13. Besides that, FL also demonstrates its potential to solve the dilemma problem of ``isolated data island" faced by companies/organizations who hesitate to share their vast volume of data samples for business concerns and privacy regulations \cite{TIST19:Yang}. Some industrial examples include biomedical data analysis in Owkin, finance and insurance data analysis in WeBank and Swiss Re, and drug discovery in MELLODDY \cite{Others19:Kairouz}, \cite{INFOCOM21:Tang}.  

The incentive issues are paramount and indispensable to FL training. FL consumes plenty of resources from participants, such as computation power, bandwidth, and private data, some of which might be constraint in scenarios like mobile networks and mobile edge computing. In addition, participants still worry about security and privacy threats in FL, where some attacks have already been found recently \cite{INFOCOM19:Wang}. All these factors hinder the participation of clients in FL without enough payback\footnote{In this paper, the payment includes cash payment, business reputation, the share of well-trained model, etc.}. Furthermore, the training performance of FL, e.g., model accuracy, training speed, will be deteriorated without sufficient training data, communication bandwidth, and computation power provided by participants \cite{ICDCS20:Zeng}. In other words, deficient participants might cause FL to malfunction in reality. Therefore, incentive mechanisms are required to inspire more clients with high-quality data and sufficient resources to engage in cooperative learning, which finally achieves the performance improvement of FL. 

Fortunately, incentive mechanism has attracted great interest and some impressive studies mushroom in the last two years. Among these results, Zhan publishes a representative survey of incentive mechanism for FL \cite{TETC21:Zhan}, and they summarize the existing studies into three categories, i.e., clients' contribution, reputation, and resources allocation. Compared with this valuable overview, we provide another survey of incentive mechanism with the distinct understanding, comprehensive taxonomy, totally innovative summary perspective, and differential insights for the future studies. These two complementary surveys provide a systemic and comprehensive summary together to interested readers.

Initially, we identify the problem of incentive scheme in FL and highlight the statement that the final goal of incentive design is to improve the training performance of FL. This is because that a few high-quality participants outperform a large number of staggers in FL. This special requirement makes the incentive issue in FL quite different from those in other scenarios. We further point out three components of incentive mechanisms, i.e., contribution evaluation, node selection, and payment allocation, and also present a novel taxonomy for further review. 

Subsequently, we explicitly summarize the existing studies in the roadmap of main techniques, which contain Shapley value, Stackelberg game, auction, contract theory, reinforcement learning, blockchain, etc. Among them, Shapley value is usually adopted for contribution evaluation, while payment allocation mostly involves Stackelberg game, auction, contract theory, and non-convex maximization. Some cutting-edge techniques such as reinforcement learning and blockchain are adopted as auxiliary mechanisms for node selection, contribution evaluation and robustness improvement. It should be noted that one technique is used for many functionalities instead of only one. In this survey, we also underline the assumptions of these studies, which is the key point of mechanism design and will definitely benefit future studies. 

By comparing the pros and cons of prominent studies from multi-dimensional views, we finally obtain some insights of opportunities and challenges in the future studies. We believe that (1) the performance improvement is the top priority of incentive mechanisms in FL. The implicit relationship between incentive mechanisms and performance improvement should be considered seriously in the design of FL; (2) some scenarios such as Mobile Edge Computing (MEC), 5G/B5G, and IoTs add special constraints to the incentive designs in FL. The incentive mechanisms should be well designed for these specific application scenarios; and (3) incentive mechanisms for cross-silo FL are neglected in current studies, and we should put more emphasis on cross-silo FL in the near future. 

The rest of this paper is organized as follows. In Section II, We present the introduction of FL, the problem statement of incentive mechanism, and its taxonomy. Then, we separately summarize the existing incentive schemes in terms of main techniques adopted in Section III. In Section IV, we compare some representative results and then point out three directions in the future study. Section V concludes the entire paper. 

\section{Federated Learning, Problem Statement, and The Taxonomy}
\subsection{Federated Learning}
\begin{figure}[tp]
\centering
\includegraphics[scale = 0.44, width =8.2cm]{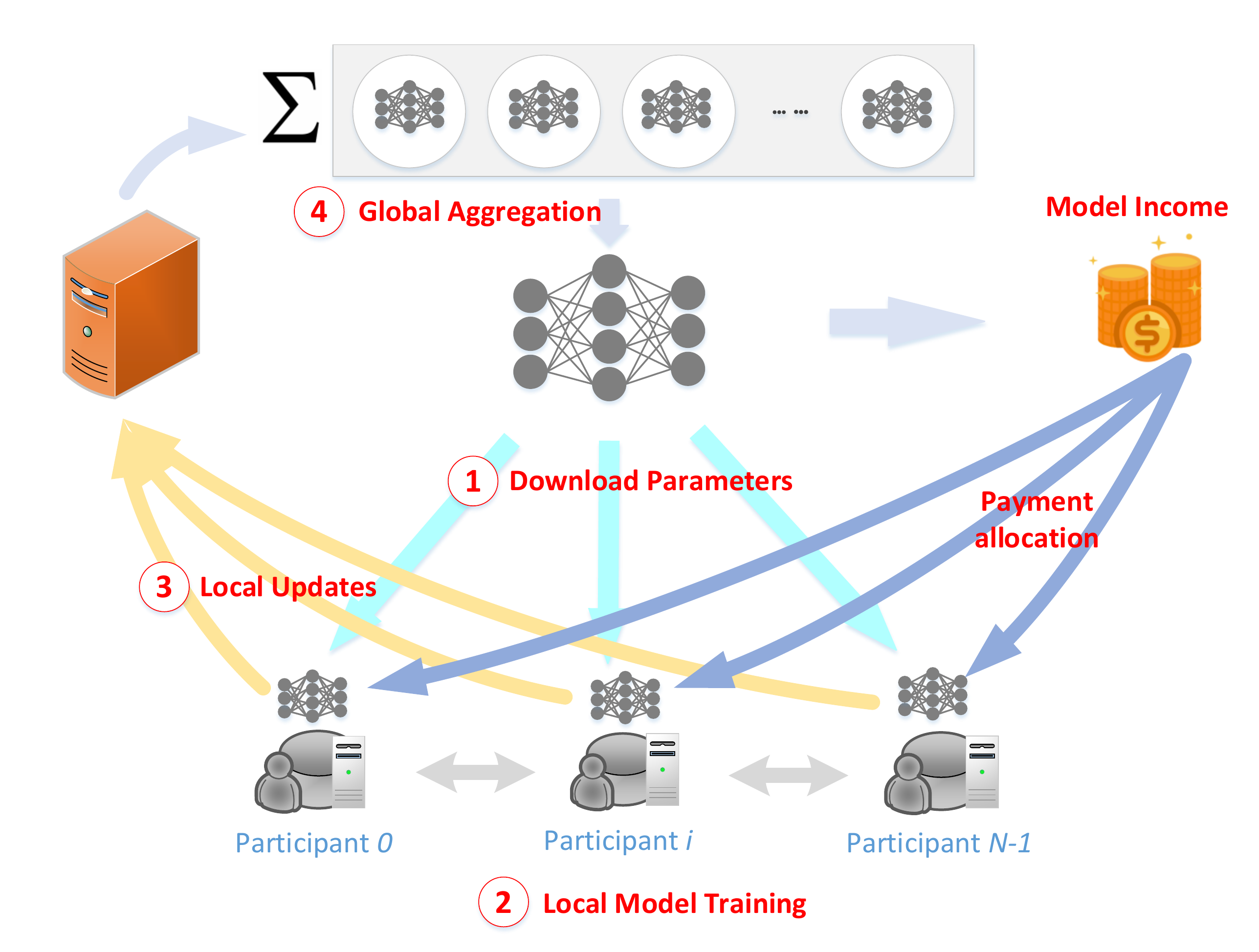}
\vspace{-0.1cm}
\caption{The system architecture of FL}
\label{SysModel}
\vspace{-0.2cm}
\end{figure}

Federated learning is a distributed training paradigm which aims at minimizing the loss function $L(\mathbf{w})$ of global model with many participants in a collaborative way. Typically, several rounds of training are involved to obtain the optimal model parameters $\mathbf{w}^*(t)$. In each round of training, the parameter server firstly distributes the global parameter $\mathbf{w}(t)$ to some participants. Then, chosen participant $i$ performs local model training. In horizontal FL, where each participant contains distinct data samples with identical features and local model, local training is to compute gradient descent as $\mathbf{w}_i(t+1) = \mathbf{w}_i(t+1) - \eta \nabla L_i(\mathbf{w}_i(t))$ with the local data set $D_i$. In vertical FL, where participants contains identical data samples with different features, data alignment and information exchange are required among participants in the local training, besides the computation of gradient descent. After finishing local training, participants upload their local parameters $\mathbf{w}_i(t+1)$ to the global server. When the global server obtains enough updates, it aggregates these parameters with algorithms like FedAgv to get a new parameter $\mathbf{w}(t+1)$ of global model. This process iterates until the global model accuracy is satisfied or the training time exceeds the predefined threshold $T$. The training procedure of FL is shown in Fig. \ref{SysModel}. In Fig. \ref{SysModel}, participants might be mobile devices, edge nodes, and IoTs devices in cross-device FL or giant companies in cross-silo FL. They provide various types of resources instead of only data, all of which are key factors to the training performance. 

\subsection{Problem Statement of Incentive Mechanisms in FL}
Most Incentive mechanisms are to inspire more qualified clients to participate in FL training with sufficient resources. In the following, we present the theoretical formulation of incentive mechanism.
\begin{myDef}[Incentive Mechanism]
(1) There exists a set of $|\mathcal{N}|$ potential candidates $\mathcal{N} = \{0, \cdots, N-1\}$, and each candidate $i$ has the multi-dimensional contribution denoted by the vector $\mathbf{q}_i$ and the type profile vector $\theta_i$. As a rational individual, it maximizes the profit as 
\begin{equation}
\pi_i=p_i-c_i(\mathbf{q}_i,\theta_i),
\label{ParticipantPI}
\end{equation} 
where $p_i$ is the payment obtained from the model owner and $c_i(\cdot)$ is the corresponding cost function. 

(2) The global server (or model owner) maximizes its profit function as 
\begin{equation*}
\pi= U(Q) - \sum p_i,
\end{equation*} 
where $U(\cdot):Q \mapsto \mathbb{R}$ is called the utility function and $Q=(\mathbf{q}_0, \cdots, \mathbf{q}_{N-1})$.

(3) The incentive mechanism is to find the optimal $Q$ in order to obtain some desirable properties or maximize some objective functions such as social welfare. Any candidate $i$ with $\mathbf{q}_i = 0$ indicates that it is not selected for FL training.  
\end{myDef}

In Definition 1, three important issues are left to be explicitly explained. One issue is related to the optimal solution $Q$ for each player\footnote{In this paper, we use the term player to indicate both participants and global server (the model owner).}. Among various solutions, Nash Equilibrium (NE) strategy is desirable since no player has incentive to choose another resource provision. The formal definition of NE is given in Definition 2. Besides the NE strategy, there exists some other types of equilibrium solutions, such as dominant strategy equilibrium, Bayesian equilibrium, and perfect Bayesian equilibrium. Interested readers refer to \cite{FLWorkshop19:Cong} for more details. 

\begin{myDef}[Nash Equilibrium]
The strategy set of all the players $(\mathbf{q}_0^{NE}, \mathbf{q}_1^{NE},\cdots, \mathbf{q}_{N-1}^{NE})$ is a Nash Equilibrium, if any participant $i$ has 
\begin{equation*}
\pi_i (\mathbf{q}_i^{NE}, \mathbf{q}_{-i}^{NE}) \ge \pi_i (\mathbf{q}_i, \mathbf{q}_{-i}^{NE}), \forall \mathbf{q}_i \ge 0,
\end{equation*} 
where $\mathbf{q}_{-i}^{NE} = (\mathbf{q}_0^{NE}, \cdots, \mathbf{q}_{i-1}^{NE}, \mathbf{q}_{i+1}^{NE}, \cdots, \mathbf{q}_{N-1}^{NE})$, and $\mathbf{q}_i \in S_i$ is one of strategies for participant $i$. 
\end{myDef}

Another issue is about the desirable properties of incentive mechanisms. Most previous incentive schemes in other scenarios target at the properties of Incentive Compatibility (IC), Individual Rationality (IR), fairness, Pareto Efficiency (PE), Collusion Resistant (CR), and Budget Balance (BB), all of which are also the focus of incentive mechanisms in FL. The illustrations of these characteristics are listed as follows.  

\begin{itemize}
    \item Incentive Compatibility (IC): An incentive mechanism has the property of IC when it is optimal for all the players to truthfully declare their contributions and cost types. In other words, reporting the counterfeit information will not benefit the revenue for the malicious player. 
    \item Individual Rationality (IR): An incentive mechanism is IR only if all the participants have non-negative profits, i.e., $\pi_i \ge 0$, for $\forall i$. IR indicates that candidates hesitate to join in FL when its payment is less than its cost. 
    \item Fairness: Some predefined fairness functions, such as contribution fairness, regret distribution fairness, and expectation fairness \cite{AIES20:Yu}, are maximized, and then the incentive mechanism achieves the property of fairness. Fairness is key to the sustainable collaboration in FL.
    \item Pareto Efficiency (PE): When the social surplus (also called social welfare), i.e., $\pi + \sum_i p_i$, is maximized, an incentive mechanism is PE, which evaluates the overall profit of federated learning. 
    \item Collusion Resistant (CR): An incentive mechanism is CR if none of subgroups of participants can obtain higher profits by conducting fraudulent activities in collusion.
    \item Budget Balance (BB): A scheme is BB iff the sum of payment for participants is no more than the budget given by the model owner or the global server. 
\end{itemize}

\begin{figure}[tp]
\centering
\includegraphics[scale = 0.44, width =8.2cm]{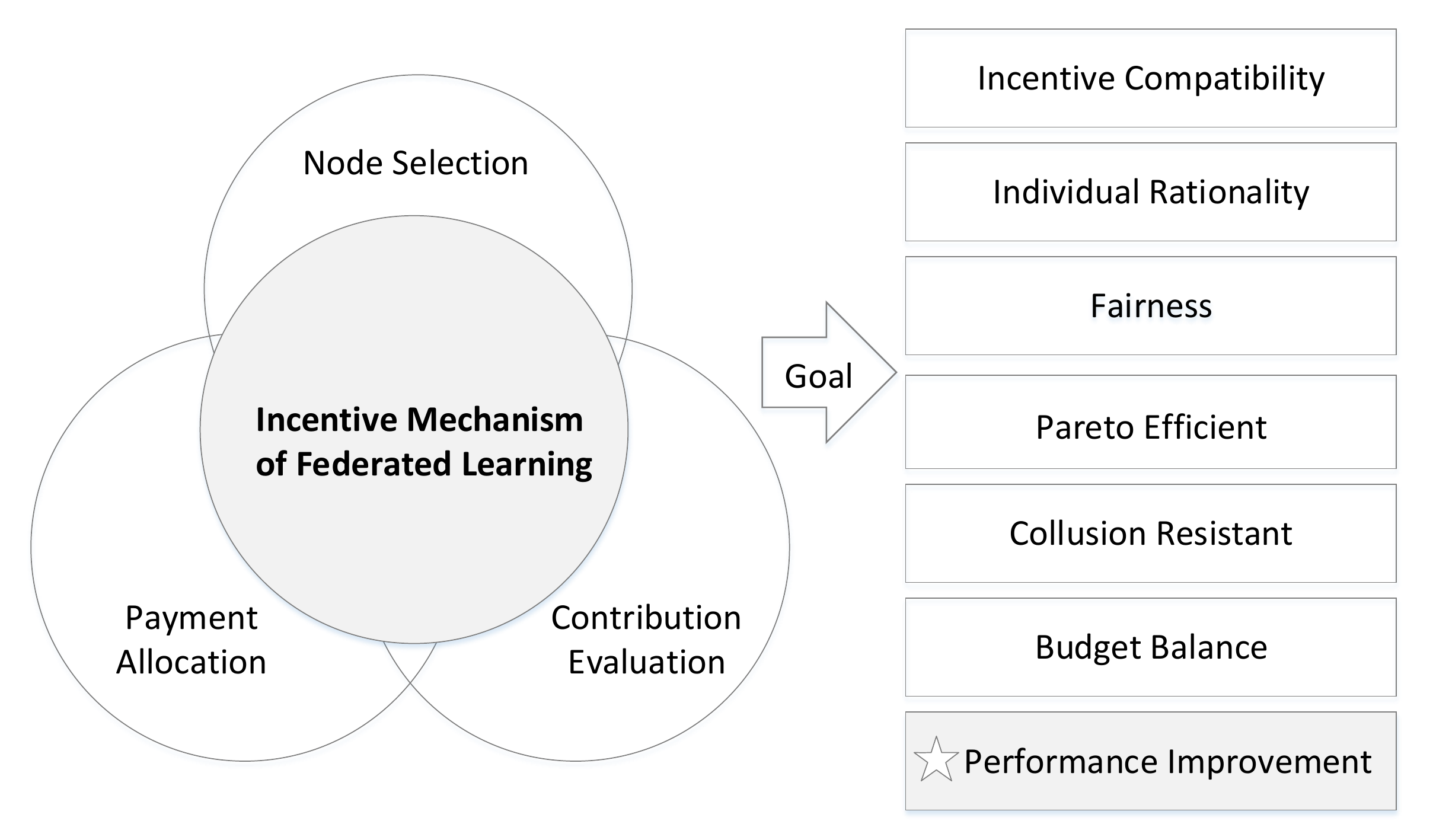}
\vspace{-0.1cm}
\caption{The framework of incentive mechanism in FL}
\label{Framework}
\vspace{-0.2cm}
\end{figure}

In FL, we argue that the incentive mechanism should have additional requirement of Performance Improvement (PI), which makes it quite different from the classic incentive mechanisms. This specific requirement stems from the phenomenon that high-quality participants outperforms many staggers in FL. In other word, inspiring a large number of participants with constraint resources and low-quality training data may negatively impact the performance of FL. Thus, incentive mechanism should achieve the property of PI in FL \cite{ICDCS20:Zeng}, \cite{ComSurvey21:Gu}. 

\begin{myRemarks}[Incentive Mechanisms of FL]
The incentive mechanisms in FL should ultimately improve the performance of FL in terms of model accuracy, training speed, communication overhead, computational costs, etc. 
\end{myRemarks}

The third issue involves the sub-components of incentive mechanism. In FL, incentive mechanism is composed of contribution evaluation, node selection, and payment allocation. Contribution evaluation endeavors to get the accurate measurement of contribution $\mathbf{q}_i$ for each participant. The simplest measurement is just the size of training data in many existing studies. Meanwhile, Shapley value is a powerful tool for contribution evaluation, which will be explicitly illustrated in Section III.A. In addition, we can find some studies of data value in other research area, and it might be considered as a ingredient of contribution evaluation. Interested readers refer to \cite{Others19:Ghorbani} for more details. Node selection is to choose a subset of qualified participants to join in FL training. In FL, the criteria of node selection not only cover the basic resources requirement but also involves the economic factors, i.e., contributing most with least cost. Payment allocation decides the payment for each chosen participant. It should be mentioned that these three components are interdependent. And a single technique, such as procurement auction, may involve several sub-components. The framework of incentive mechanism is shown in Fig. \ref{Framework}. 

\subsection{The Taxonomy of Incentive Mechanisms} 
\begin{figure}[tp]
\centering
\includegraphics[scale = 0.36]{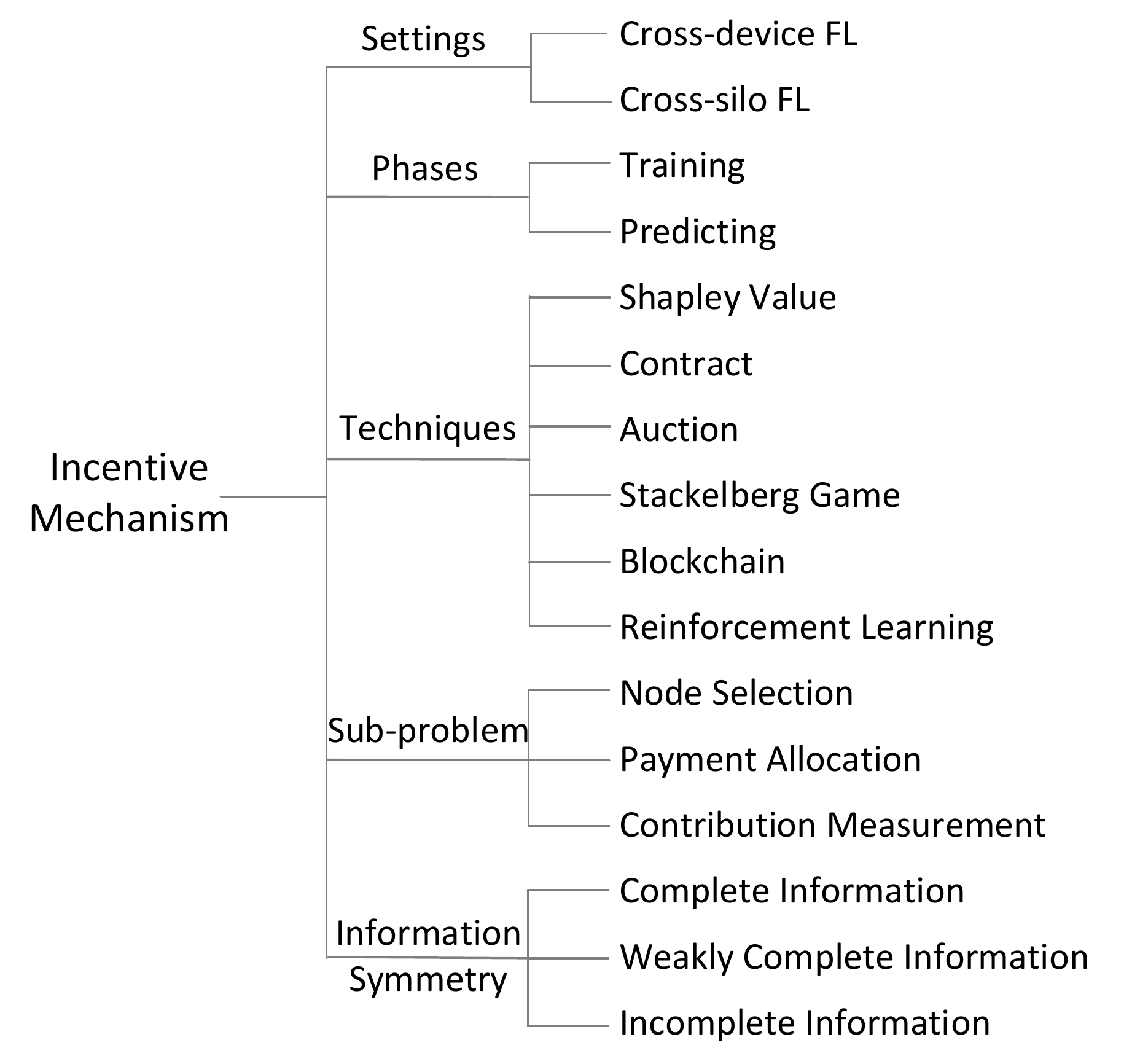}
\vspace{-0.1cm}
\caption{The taxonomy of incentive mechanism in FL}
\label{Taxonomy}
\vspace{-0.2cm}
\end{figure}

The existing studies of incentive mechanisms can be categorized in terms of application settings, the FL phase, main techniques, sub-problems, and the assumption of information symmetry. Fig. \ref{Taxonomy} presents our proposed taxonomy of incentive schemes in FL. 

\begin{itemize}
\item Application Settings: FL can be applied to both cross-device scenarios and cross-silo scenarios. In cross-device FL, a large number (about $10^2$ to $10^{10}$) of vulnerable mobile node, IoTs devices, or smart edge with non-IID training data and constraint resources dynamically and collaboratively train a global model for the third party such as the application service provider. On the contrary, less than 100 giant organizations/companies, such as medical hospitals and financial companies, train a shared model with sufficient data and geo-distributed data centers in cross-silo settings. These two scenarios are quite distinct in a series of aspects. More comparisons can be found in \cite{Others19:Kairouz}. Currently, most studies focus on cross-devices cases \cite{INFOCOM21:Ding}, \cite{IoTJ20:Lim}, \cite{INFOCOM21:Tang}, while a few studies such as \cite{BigDate:Song} and \cite{INFOCOM21:Deng} target at cross-silo FL. 
\item FL Phase: The entire process of FL includes model training and model prediction. The training phase targets at obtaining high-quality global model with efficient training performance, while model prediction focus on truthful response of prediction results with many well-trained models of participants. The design goals and assumptions vary these two phases, which leads to distinct incentive schemes \cite{INFOCOM21:Weng}.
\item Main Techniques: The techniques adopted in incentive mechanism include Stackelberg game, auction, contract, Shapley value, blockchain, and reinforcement learning. Stackelberg game \cite{INL19:Sarikaya}, \cite{IoTJ20:Zhan}, \cite{Others18:Feng}, auction \cite{ICDCS20:Zeng}, \cite{TMC20:Jiao}, \cite{WCNC20:Le}, and contract \cite{Others19:Kang}, \cite{WiOpt20:Ding} are mainly employed by node selection and payment allocation, while Shapley value \cite{BigData19:Wang}, \cite{BigDate:Song} is used for contribution measurement. Both blockchain \cite{Others20:LiuYuan} and reinforcement learning \cite{TMC20:Jiao}, \cite{INFOCOM20:Wang} are auxiliary techniques to improve the performance and robust incentive schemes. In Section III, we summarize the existing studies with a roadmap of main techniques.
\item Sub-problems: As shown in the above subsection, the sub-problems of incentive scheme include contribution evaluation \cite{GLOBECOM20:Nishio}, node selection \cite{INFOCOM20:Wang}, and payment allocation \cite{Others20:LiuYuan}. It should be noted that some studies solve several sub-problems of incentive mechanism simultaneously instead of only a single sub-problem. 
\item Information Symmetry: According to the assumption of information symmetry, the existing schemes can be categorized into incentive schemes with incomplete information \cite{Others20:Lim}, \cite{INFOCOM2015:Zhao}, weakly complete information \cite{WiOpt20:Ding}, and complete information \cite{INL19:Sarikaya}, \cite{GLOBECOM19:Shashi}, \cite{Others19:Khan}.
\end{itemize} 

\section{Incentive Mechanism of FL}
In this section, we summarize the state-of-art incentive mechanisms in a technical way. Specifically, we separately present the schemes with Shapley value, Stackelberg game, auction, contract, reinforcement learning, blockchain, and other techniques.

\subsection{Shapley Value}
Shapley value originated in cooperative game theory is widely adopted by incentive mechanisms of FL, especially for the contribution evaluation and profit allocation \cite{BigData19:Wang}, \cite{BigDate:Song}. Since there does not exist an explicit linear relation between data size of individual participant and accuracy of global model, it is unreasonable to use the metric of data size to directly evaluate its contribution for each participant. For instance, providing lots of low-quality data sample may not help building a global model \cite{BigData19:Wang}. Moreover, the situation becomes much complicated when multi-dimensional resource provision is considered in FL \cite{ICDCS20:Zeng}. Based on these observations, Shapley value is employed to evaluate the contribution from the view of utility and impacts on the collaborative training. 

The core idea of Shapley value is to calculate the weighted average of marginal contribution as Shapley value. Let function $v(\mathcal{S})$, where $\mathcal{S} \subseteq \mathcal{N}$, be the utility of model collaboratively trained by the subset $\mathcal{S}$. The Shapley value $\varphi(i)$ of participant $i$ is given as
\begin{equation}
\varphi(i) = \sum_{\mathcal{S} \subseteq \mathcal{N}\backslash \{i\}} \frac{|\mathcal{S}|!(N-|\mathcal{S}|-1)!}{N!}\{v(\mathcal{S} \cup {i}) - v(\mathcal{S})\}.
\label{ShapleyValue}
\end{equation} 
From Eq. \ref{ShapleyValue}, we can find that $v(\mathcal{S} \cup {i}) - v(\mathcal{S})$ is marginal contribution when participant $i$ joins in the federated training with the subset $\mathcal{S}$. In the following, we present a toy example to demonstrate the calculation of Shapley value. 
\begin{myDemo} Suppose $\mathcal{N} = \{0,1,2\}$. The utilities of different coalitions are listed as follows: 
\begin{eqnarray*}
v(\emptyset) = 0 && v(\{0\}) = 5 \\
v(\{1\}) = 10 && v(\{2\}) = 15 \\
v(\{0, 1\}) = 30  && v(\{0, 2\}) = 40 \\
v(\{1, 2\}) = 60 && v(\{0, 1, 2\}) = 100
\end{eqnarray*} 
The calculation of each participant is shown in the following table. In the first line ($0 \gets 1 \gets 2$), the marginal contributions of node $\{1\}$ and $\{2\}$ are separately $v(\{0, 1\}) - v(\{0\}) = 25$ and $v(\{0, 1, 2\}) - v(\{0, 1\}) = 70$. Similarly, we can get all the marginal contribution for other lines. Finally, the Shapley value for each node can be obtained as the average value of corresponding column.  
\begin{table}[!htp]
\centering
\begin{spacing}{1}
\begin{tabular}{cccc}
\hline
  & $\quad 0 \quad$ & $\quad 1 \quad$ & $\quad 2 \quad$\\
\hline 
$\quad 0 \gets 1 \gets 2 \quad$& 5 & 25 & 70\\
$\quad 0 \gets 2 \gets 1 \quad$& 5 & 60 & 35\\
$\quad 1 \gets 0 \gets 2 \quad$& 20 & 10 & 70\\
$\quad 1 \gets 2 \gets 0 \quad$& 40 & 10 & 50\\
$\quad 2 \gets 0 \gets 1 \quad$& 25 & 60 & 15\\
$\quad 2 \gets 1 \gets 0 \quad$& 40 & 45 & 15\\
\hline
sum & 135 & 210 & 255\\
\hline
$\varphi(i)$ & 22.5 & 35 & 42.5\\
\hline
\end{tabular}
\end{spacing}
\end{table}
\end{myDemo}

However, Shapley value is extremely computationally expensive. The computation complexity of Shapley value is $\mathcal{O}(N!)$, which performs exponential number of utility calculation \cite{Others19:Ghorbani}. For this computation problem, a series of studies in data science community have been published for approximating Shapley value with sampling techniques such as Monte Carlo, group-testing based approach, probabilistic estimation, and proper information sharing \cite{AISTATS19:Jia}. Interested readers refer to \cite{AISTATS19:Jia} and \cite{Others19:Ghorbani} for more details.

When Shapley value is adopted by FL, the computation optimization is also an interesting topic. In \cite{BigDate:Song}, Song et al. consider the characteristics of FL and propose two gradient-based methods to efficiently calculate model utilities for different combinations of enterprise participants. The first method called one-round reconstruction based algorithm collects all the local gradients in FL training and reconstructs all the models after training to compute marginal contributions. The other method utilize the additivity of Shapley value to calculate contributions in each training round and then aggregate them to get the final result. These two methods trade additional storage of local gradients for the calculation of marginal contributions without the need of retraining each model.      

Liu proposes a decentralized approximation approach with blockchain system for Shapley value in FL \cite{Others20:LiuYuan}. In their framework FedCoin, each miner independently and competitively computes the approximated Shapley value with sampling technique, and the miner whose result is the nearest to the average of all the minders is chosen as the winner. The winner generates a new block and records the payment allocation according to the average Shapley value in an immutable manner. FedCoin demonstrates its capability to calculate Shapley value with an upper bound on the computational resource for reaching consensus. In a nutshell, how to efficiently apply Shapley value in resources-constraint FL is a troublesome problem left for the future work.

Besides the computational issue, data privacy is another potential problem for Shapley value in FL. The direct adoption of Shapley value might reveal the protected feature value or data sample distribution during the computation of marginal utility for each coalition. In \cite{BigData19:Wang}, Wang et al. firstly point out this threat in vertical FL and adopt a variant version called Shapley group value to measure the utility of a feature subset without revealing the details of any private feature in vertical FL. In detail, they combine some private features as a united federated feature and compute the Shapley group value for this federated feature in a two-participants case. When the number of participant increases, the computation becomes an intractable issue. 

In FL, Shapley value might has a broad application such as contribution measurement of participant \cite{BigData19:Wang}, \cite{BigDate:Song}, \cite{Others20:LiuYuan}, participant behaviour analysis \cite{IJCAI20:Ng}, and feature selection \cite{NIPS17:Lundberg}. Especially, Ng et al. implement multiple payoff-sharing schemes (including Shapley value) in a visual tool of a multi-player game to study how FL participants act under different incentive schemes through crowdsourcing in \cite{IJCAI20:Ng}.

\subsection{Stackelberg Game} 
Stackelberg game is a sequential model of game theory commonly used to formulate the interactions between different players in the sell or procurement of common products, making it more appropriate for the incentive design in FL \cite{INL19:Sarikaya}, \cite{IoTJ20:Zhan}, \cite{INFOCOM18:Yu}. In Stackelberg game, there exist two categories of players, i.e., leader and follower. The leader optimizes its profit as $\max_\Phi \pi(q_1, \cdots, q_N, \Phi)$ by considering the expected reactions of followers and moves first to declare its decision $\Phi$. Subsequently, follower observes the action of leader, optimizes its own profits, and responses $\pi_i (i \in \mathcal{N})$. Usually, the leader or follower is a collection of players instead of one participant, and a group of players perform Cournot game simultaneously. In Stackelberg game, the NE solution can be obtained via backward induction. In details, we first get the optimal responses $q_i = g_i(\Phi) = \arg \max \pi_i(q_1, \cdots, q_N, \Phi)$ by setting first-order derivative with 0, and then substitute them into the objective function $\max \pi(g_1(\Phi), \cdots, g_N(\Phi), \Phi)$. In this way, the NE solution $(\Phi^{NE}, q_1^{NE}, \cdots, q_N^{NE})$ can be computed. 

Some major issues require to be tackled when Stackelberg model is used in FL settings. (1) \emph{Who are the leader and follower?} Most of studies consider the FL ecosystem as a monopoly market, where the single leader is the model owner or the cloud server and the followers are multiple participants performing Cournot game in the second stage. These studies are conducted in scenarios of mobile edge computing \cite{GLOBECOM19:Shashi}, distributed coded machine learning \cite{INFOCOM18:Yu}, IoTs \cite{GLOBECOM20:Hu}, etc. On the contrary, Feng argues that model owner uses the learning service provided by mobile devices, and the model owner acts as a single follower in the lower level of Stackelberg game with multiple leaders \cite{Others18:Feng}. In this mobile scenario, many mobile leaders simultaneously perform Cournot game in the first stage of Stackelberg game, while the single follower responses them with the total requirement of training data. The differences of these two categories of studies lie in the market assumption of FL scenarios, and their justifications are left for future studies. 

(2) \emph{What is the ``common product'' in FL?} Intuitively, the product is training data samples evaluated by its size, and model owner trade rewards for training data in \cite{Others18:Feng}, \cite{IoTJ20:Zhan}. The second type of product is computation power for local training. In \cite{INL19:Sarikaya}, Sarikaya employs Stackelberg game model to motivate the CPU provisions of multiple workers to reduce the local training time with given budget of the leader in FL with a fully-synchronous SGD. Similarly, Ding considers a multi-dimensional product distinguished by the computation speed, the start-up computation time and the cost in distributed coded machine learning \cite{INFOCOM21:Ding}. In order to accelerate the computation of NE solution, they ingeniously transform the multi-dimensional feature into a single metric. Other impressive products include local training accuracy and data privacy. In \cite{GLOBECOM19:Shashi}, the leader decides uniform reward rate (e.g., $\$/$Accuracy Level), and the follower participants optimize their utility functions with the variable of accuracy level which can be further used in user selection. In \cite{GLOBECOM20:Hu}, Yu argues that clients should be compensated for their privacy loss in FL and adopt the Stackelberg game approach to inspire more clients contributing private data. The measurement of privacy utilizes the $\rho$-zero-concentrated differential privacy ($\rho-z$CDP), and the payment allocation is proportional to the privacy budget of each client.

(3) \emph{How to obtain NE solution with complete information and incomplete information?} The NE solution can be theoretically computed with backward induction in complete information scenario \cite{Others19:Khan}. For the practical case with incomplete information and stochastic information, Ding finds that the computation complexity of NE solution is increased with additional $N(N-1)$ IC constraints \cite{INFOCOM21:Ding}. They not only present the optimal strategies but also study the impacts of workers' private information on the NE solution. In \cite{IoTJ20:Zhan}, Zhan believes that the assumption of shared decision information among participants does not hold in realistic IoT applications and the inaccurate contribution evaluation negatively impacts the incentive mechanism design. Based on these observations, they propose a Deep Reinforcement Learning (DRL) based scheme to dynamically adjust players' strategies and optimize their profits in the scenarios with incomplete information and ambiguous contribution measurement. In the proposed DRL scheme of Proximal Policy Optimization (PPO), the model owner separately uses the actor network and critic network to maintain its policy and estimated value function. In addition, each edge node trains a complicated DRL model with high computation overhead and uses an offline training mode to learn its optimal strategies in a simulation environment. In a nutshell, DRL seems to be a promising technique for computing NE solution in game theory.  

(4) \emph{Does the proposed model improve the performance of FL?} Two impressive studies correlate the incentive design with the performance of FL model. In \cite{INFOCOM21:Ding}, Ding shows that the optimal recovery threshold should be linearly proportional to the total participator number when using MDS codes. In cooperative relay networks, authors balance the tradeoff between the provision of training data and the participation of relay services \cite{Others18:Feng}. Also, they significantly reduce the congestion in the communication for both the model owner and mobile devices. In sum, we believe that the performance-aware incentive schemes are more appreciated by FL.   

\subsection{Auction}
Auction is another efficient mathematical tool for pricing, task allocation, node selection, etc., and it has been extensively used in radio frequency spectrum allocation, advertising, and bandwidth allocation in computer communities \cite{ComSurveys11:Parsons}. In auction, there exist two types of players, i.e., the auctioneer and bidders. The single auctioneer served by the global model owner or cloud server coordinates the process of auction, while bidders are participants who response the auctioneer with various local resources and their bids. The detailed process of auction can be generalized as bid announcement, bid collection, winner determination, and some other auxiliary procedures such as clearing price, information revealing, etc. In each step, it may contains some specific methods such as first-price payment, second-price payment, etc., which generates a veritable menageries of auctions \cite{ICML19:Duetting}. Interested readers refer to \cite{ComSurveys11:Parsons} for deeper technical understanding. In sum, auction allows participants to actively report their true type, which makes it more impressive than other game tools.  

Currently, \cite{ICDCS20:Zeng}, \cite{Others20:Cong}, \cite{TMC20:Jiao}, \cite{WCNC20:Le}, and \cite{INFOCOM21:Deng} are the representative studies with auction-based incentive scheme in FL. Specifically, Zeng proposes a lightweight and multi-dimensional incentive scheme FMore with procurement auction for FL in mobile edge computing scenarios \cite{ICDCS20:Zeng}. In the phase of bid announcement, the auctioneer advertises a public scoring function and its resource requirement. After receiving this advertisement, participants decide the type of resource provision and the corresponding price by separately maximizing their own profits, and then send these bids back to the auctioneer. With these bids, the remote server chooses $K$ winners based on the computed score and determines their payments. In \cite{INFOCOM21:Deng}, Deng proposes a quality-aware auction scheme with the similar process in a multi-task learning scenario. One innovative point is that they formulate the winner selection problem as an NP-hard Learning Quality Maximization (LQM) problem and devise a greedy algorithm to perform real-time task allocation and payment distribution based on Myerson's theorem. Actually, most winner selection and payment allocation problems are computationally intractable, and thus randomized auction is adopted as a simplified and effective solution in FL. For example, Le adopts the framework of randomized auction to obtain the approximate solution to the knapsack problem of social cost minimization. The proposed framework contains three components: 1) an greedy algorithm which achieves $ln C$-approximation efficiency, 2) convex decomposition, and 3) VCG-based payment allocation \cite{WCNC20:Le}. Similarly, Jiao proposes a reverse multi-dimensional auction (RMA) mechanism which follows a randomized and greedy method to choose participants and decide the payments in wireless FL scenarios \cite{TMC20:Jiao}. Finally, Cong proposes an impressive VCG-based incentive mechanism Fair-VCG which not only maximizes social welfare but also minimizes unfairness of the federation in \cite{Others20:Cong}. In Fair-VCG, the proposed approach involves three steps, i.e., the computation of data acceptance vector, the computation of VCG payment vector, and the computation of adjusted payment vector, and its computation complexity is undoubtedly NP-hard. 

From these studies, we could find some common features of auction-based incentive schemes. (1) \emph{Most of the proposed schemes are multi-dimensional}. In \cite{ICDCS20:Zeng}, the server advertises its requirements of various resources, such as data size, computation power, communication bandwidth, etc. In the step of winner determination, authors adopt functions like a Cobb-Douglas function, a perfect complementary function, or a perfect substitution function to transform multi-attributes into a scalar, and evaluate this scalar value by a public scoring function. In \cite{TMC20:Jiao}, Jiao considers data size, data distribution evaluated by the metric of Earth Mover's Distance (EMD), and requested wireless channel in the incentive mechanism. Instead of incorporating them into one metric, RMA treats them separately in the random and greedy search of winners. In \cite{WCNC20:Le}, authors consider the uplink transmission power and the computation resource (CPU cycle frequency) with a given subchannel bundle for mobile participants in cellular networks.

(2) \emph{The performance improvement of FL is one of main goals of incentive design}. In \cite{ICDCS20:Zeng}, Zeng argues that the proposed scheme FMore is able to inspire high-quality nodes with sufficient resources to join in FL training and further improve the performance of FL. Results show that FMore speeds up FL training via reducing training rounds by 51.3\% on average and improves the model accuracy by 28\% for the tested CNN and LSTM models. Similarly, Deng proposes the quality-aware FL framework FAIR which integrates incentive mechanisms with local learning quality estimation and model aggregation in \cite{INFOCOM21:Deng}. Interestingly, they adopt an exponential forgetting function to generate local quality estimation for node selection, while the proposed model aggregation considers both the data size and estimated quality simultaneously. Simulation results demonstrate that the model accuracy with FAIR outperforms the knapsack greedy mechanism when the percentage of mislabeled data ranges from 20\% to 80\%. 

(3) \emph{Machine learning is smartly adopted as a auxiliary technique for auction}. In \cite{TMC20:Jiao}, Jiao proposes a deep reinforcement learning-based auction (DRLA) mechanism to improve the social welfare and simultaneously ensure IC and IR. In addition, they also use graph neural networks to generate the embeddings of conflict relationship between participants and then apply these embeddings into the DRLA to automatically determine service allocation and payment. Neural networks can also be adopted to solve the functional optimization problem, since they can approximate any continuous function with the predefined precision \cite{Others20:Cong}. In Fair-VCG scheme, the unsupervised and composite neural networks of $2n$ small networks are developed to approximate two continuous and increasing functions which are further used for winner determination and payment allocation. 

\subsection{Contract Theory} 
Contract theory is to study how players construct and develop optimal agreements with conflicting interests and different levels of information. Among various types of contract, public procurement contract is widely adopted for the incentive mechanism design. In the public procurement contract, the server offers a menu of contracts to participants, without being informed about the private cost of participants at the time contracts are written, and each participant proactively pick the option that was designed for its type. The public procurement contract embodies the self-revealing property which could elicit the optimal provisions from participants with the presence of information asymmetry. On the contrary, the performance of information revelation is determined by the granularity of contract lists, since participants can only choose a contract. 

In FL, the existing studies can be categorized into two groups, i.e., contracts with different assumptions of information asymmetry and multi-dimensional contracts \cite{IoTJ20:Lim}. In the first groups of studies, authors in \cite{Others19:Kang} initially identify the information asymmetry between the task publisher and participants, since the private information of data size and various resources are unknown for the task publisher like the base station. They adopt contract theory to map the contributed resources into a appropriate rewards via contract lists. In details, they consider the local computing power as the decision variable for participant and assume a strongly incomplete information scenario where they only have the knowledge of the probability that a participant belongs to a certain type. Formally, the base station optimizes its utility and publishes a contract list $(R(f_n), f_n)$, where $R(f_n)$ is the reward of the computation resources of type-$n$ participant. After receiving the contract list, a participant chooses its own true type $n$ which can maximize its private utility. For NE solution, the monotonicity constraint is added with IC and IR to solve the optimization problem for the task publisher. They further extend this study in \cite{Others19:Kang} by combing contract theory with reputation and blockchain in \cite{IoTJ:Kang}. They assume that adversarial participant might perform malicious parameter update to deteriorate the performance of global model, adopt the metric of reputation calculated with a multi-weight subjective logic model to measure the reliability and trustworthiness of mobile nodes, and then select some high-reputation participants as candidates. In this study, the consortium blockchain with properties of tamper-proof and non-repudiation is leveraged to secure the reputation management as well. 

Another impressive incentive mechanism in the first category theoretically studies multi-dimensional contract for three scenarios, i.e. complete informational scenario, weakly incomplete information scenario, and strongly incomplete information scenario \cite{WiOpt20:Ding}. In complete information case where the server knows the true type of each participant, the contract design is to minimize the total sever cost with IR constraint. In the scenario of weakly incomplete information, they consider the same objective function with both IR and IR constraints, since the server does not know the type information of user and cannot force it to accept a certain contract. In strongly incomplete information scenario, authors minimize the expected server cost due to the uncertainty of user type. The objective optimization function is similar to that in weakly incomplete information scenario excepted the expectation part. Since the optimization problem in this case is not necessarily convex, authors proposed a new contract structure called two-part uniform contract, the performance of which approaches to the optimal solution of original optimization problem in this scenario.

The other group of studies relate to multi-dimensional resources provisions in FL. In \cite{Others20:Lim}, Lim considers multiple resources of heterogeneity of UAVs and leverages on multi-dimensional contract theory to design a trustworthy incentive mechanism for the UAV-enable Internet-of-Vehicles (IoV) scenario. In this scheme, a participant is categorized into a multi-dimensional type-$(x,y,z,q)$, where $x$ is traversal cost type, $y$ is sensing cost type, $z$ is computation cost type, and $z$ is transmission cost type. The proposed contract design involves two-step procedure, i.e., multi-dimensional contract design and traversal cost compensation. The first step converts sensing cost and computation cost into a single-dimensional contract design problem, while the second step adds a fixed compensation to derive the final contract bundles by considering additional traversal cost and transmission cost. In \cite{WiOpt20:Ding}, Ding considers both data size and communication time as their decision variables in the contract design. They argue that selected participants should satisfy the requirement of communication time and then the incentive design is simplified into a single-dimensional contract problem. In details, they first compute the optimal rewards for any given data size, and then substitute rewards with aforementioned results to derive the optimal data size and maximum communication time. 
\subsection{Reinforcement Learning}

Reinforcement Learning (RL) is a fascinating deep learning technique to approach to the optimal solution in successive decision-making scenarios, where an agent repeatedly observes the environment, performs its actions to maximize its targets, and gets the response (usually called rewards) from the environment \cite{SPM17:Arulkumaran}. It is quite appropriate for the incentive design in FL. In FL training, the model owner considered as an agent performs the action of node selection or payment allocation to elicit high-quality participants to join in the training of FL. The agent iteratively makes decisions by trial and error and get the responses of participants (considered as rewards) to achieve the optimal training performance. From this description, we can easily find that the incentive process can be modeled by RL \cite{Others17:Hessel}. Moreover, most optimization problems in incentive design are NP-hard, and the large number of participants further aggravate this problem. Both indicate that it is almost impossible to get NE solutions, and we need an approximation technique to derive solutions for all the parties. In sum, reinforcement learning can be innovatively applied for the incentive design in FL. 

The existing studies of incentive schemes with RL can be classified into schemes with discrete action space and continuous action space. As an impressive work with discrete action set, \cite{INFOCOM20:Wang} uses the technique of double Deep Q Network (DQN) to select nodes from all the candidates to improve the performance of FL by counterbalancing the data distribution bias. In \cite{INFOCOM20:Wang}, the DQN agent calculates Q function of multiple devices, and then selects a predefined number of devices according to the Q value. Wang further replaces DQN with double DQN to make the estimation of the action-value function at the agent more stable. In \cite{TMC20:Jiao}, Jiao also employs DQN to truthful randomized multi-dimensional auction to improve the social welfare. Specifically, the agent is to select winners from finite candidates by considering channel conflicts and local data distribution, and the reward is the social welfare improvement contributed by the chosen participant. The proposed method outperforms traditional reverse multi-dimensional auction in terms of social welfare while guaranteeing the properties of truthfulness and IR.  

The other category of studies with continuous decision space mainly concentrate on payment allocation in FL. In \cite{IoTJ20:Zhan}, authors applied the impressive PPO algorithm to compute the payment of participants in Stackelberg game with incomplete information. PPO enables an agent without prior information of other players to obtain the approximate solution of NE strategy. In \cite{Networks20:Zhan}, Zhan reviews three challenges of incentive mechanisms and points out that computational resources allocation is key to a successful FL. Then, they focus on the performance of federated learning systems in real-world settings, and use PPO algorithm to design an effective incentive mechanism to maximize system performance and minimize costs simultaneously. 

\subsection{Others}
Boosted by Bitcoin, blockchain can be considered as a decentralized digital ledger in a P2P networks, where a copy of append-only ledger with digitally signed and encrypted transaction is maintained by each participant. Blockchain provides the features of robust and tamper-proof to FL, and thus some studies adopt it with incentive schemes to provide secure or privacy-preserving FL. In \cite{TDSC19:Weng}, Weng points out two serious problems of security threats from dishonest behaviors and the deficiency of incentive component in FL. For these problems, they propose a secure and decentralized framework DeepChain based on blockchain-based incentive mechanism and cryptographic primitives to provide data confidentiality, computation auditability, and incentives for parties in collaborative FL training. Similarly, Bao adopts the technique of blockchain to provide auditable federated learning with trust and incentive attributes \cite{BigCom2019:Bao}. \cite{BigData19:Toyoda} is another impressive work to propose an economic incentive approach for FL on a public blockchain. In \cite{BigData19:Toyoda}, workers in a given round choose the model updates submitted by workers in previous round and use these model parameters to train their own models with local data. Each worker votes top $k$ previous models, and the smart contract calculates the vote count of worker in previous round. The reward is distributed according to the vote count. In \cite{ICDMW18:Lu}, a commitment scheme is proposed to prevent malicious participants from simply copying and reporting other's output, and an incentive strategic game is provided to inspire participants to behave correctly.

Besides blockchain, some other impressive techniques are adopted in contribution evaluation \cite{BigData19:Wang}, \cite{GLOBECOM20:Nishio}, \cite{Others20:Cai}, node selection \cite{Others18:Nishio}, \cite{Others20:Liu}, cross-silo FL \cite{INFOCOM21:Tang}, the incentive of predication phase \cite{INFOCOM21:Weng}, fairness-aware and sustainable incentive scheme \cite{AIES20:Yu}, \cite{IIS:Yu}, etc. In cross-silo setting, Tang firstly proposes an incentive scheme from a public goods perspective and formulate this problem as a social welfare maximization problem, whose objective function is non-convex \cite{INFOCOM21:Tang}. They argue that the resources of organizations are non-excludable public goods. The proposed scheme satisfies the properties of IR and budget balance. In order to get the NE strategies, they also propose a distributed algorithm with perfect information to obtain the NE solutions. In \cite{INFOCOM21:Weng}, Weng first studies the incentive issue in the predication phase of FL, where the prediction accuracy and privacy are their top priority. Specifically, they adopt Bayesian game theory to inspire participants to truthfully make prediction for charged users, and use divergence-based Bayesian Truth Serum method to fairly reward the participants according to their truthfulness of prediction. They also employ truth discovery algorithm and TEEs to boost the accuracy of prediction and protect the model privacy. The proposed framework and incentive scheme is IR, budge balance, and the performance improvement. 

\begin{table*}[tp]
\centering
\caption{The Comparison of Some Prominent Incentive Mechanisms of FL}
\begin{spacing}{1.1}
\begin{tabular}{ccccccccccc}
\hline
  & Main Techniques & Functionality & IC & IR & PE & CR & PI & Fairness & BB & CC \\
\hline 
\cite{INFOCOM21:Tang} & Non-convex Optimization & PA & & $\surd$ & & & & & $\surd$ &\\
\cite{ICDCS20:Zeng} & Multi-dimensional Auction & NS + PA & $\surd$ & $\surd$ & $\surd$ & & $\surd$ & & & Efficient\\
\cite{Others20:Cong} & Auction + NN & NS + PA & $\surd$ & $\surd$ & $\surd$ & & & $\surd$ & $\surd$ & Large \\
\cite{INFOCOM21:Ding} & Multi-dimensional SG & NS + PA & $\surd$ & $\surd$ & & & $\surd$ & & & Efficient\\
\cite{IoTJ20:Lim} & Coalition Game + Contract & NS + PA & $\surd$ & $\surd$ & & &\\
\cite{Others19:Kang} & Contract & NS + PA & $\surd$ & $\surd$ &\\
\cite{IoTJ20:Zhan} & SG +DRL & NS + PA & & $\surd$ & & & & & & Large \\
\cite{TMC20:Jiao} & Auction+DRL & NS + PA & $\surd$ & $\surd$ & $\surd$ & & & & &\\
\cite{WCNC20:Le} & Randomized Auction & NS + PA & $\surd$ & $\surd$ & $\surd$ & & & & & Efficient\\
\cite{WiOpt20:Ding} & Contract & NS + PA & $\surd$ & $\surd$ & &\\
\cite{BigData19:Wang} & Shapley Group Value & CM & & & & & &$\surd$ & \\
\cite{INFOCOM20:Wang} & DRL & NS & & & & & $\surd$ & & & Large\\
\cite{Others18:Nishio} & Greedy Algorithm & NS & & & & & $\surd$ & & & \\
\cite{Others20:Lim} & Multi-dimensional Contract & NS + PA & $\surd$ & $\surd$ & & \\
\cite{TDSC19:Weng} & Blockchain & NS + PA & $\surd$ & $\surd$ & & $\surd$ & & &  & Large  \\
\cite{INFOCOM21:Weng} & Bayesian Game theory & CM + PA & & $\surd$ & & & $\surd$ & $\surd$ & $\surd$ & \\
\cite{INFOCOM21:Deng} & Auction & CM + PA + NS & $\surd$ & $\surd$ &  & & $\surd$ & $\surd$ & $\surd$ & Efficient \\ 
\hline
\end{tabular}
\end{spacing}
\label{ComTable}
\end{table*}

For the fairness-aware and sustainable incentive scheme, Yu studies the temporary mismatch issue between contributions and rewards in FL, and propose the Federated Learning Incentivizer (FLI) payoff-sharing scheme. FLI dynamically divides a given budget in a context-aware manner by jointly maximizing the collective utility while minimizing the inequality among the data owners. The objective function is value-minus-regret drift optimization problem, which achieves contribution fairness, regret distribution fairness, and expectation fairness simultaneously. FLI is demonstrated to be able to produce near-optimal collective utility while limiting data owner's regret \cite{IIS:Yu}, \cite{AIES20:Yu}. In \cite{Others20:Chen}, Chen studies the mechanism design of incentive in multi-party machine learning. They assume a distinct setting called interdependent value with type-dependent action spaces, where agents can never over-report its type. They also provide the optimal truthful mechanism in the quasi-monotone utility setting, and give the necessary and sufficient conditions for truthful mechanisms in the most general case. In \cite{Others20:Liu}, authors adopt the technique of information elicitation to theoretically study the scoring rule for truthful reporting of local parameters with Bayesian Nash Equilibrium strategies in two cases, i.e., elicitation with verification and elicitation without verification. 

\section{Comparisons and Future Studies}
\subsection{Comparisons and Analysis}
In this section, we compare some prominent studies from the aspects of design goals and their main functionality, shown in Table. \ref{ComTable}\footnote{Stackelberg Game (SG), Node Selection (NS), Payoff Allocation (PA), Contribution Measurement (CM), Pareto Efficient (PE), Collusion Resistant (CR), Performance Improvement (PI), Budget Balance (BB), Computational Cost (CC)}. From these comparisons, we can derive the following conclusions:  
\begin{itemize}
    \item All the node selection schemes target at the performance improvement of FL, while some recent incentive schemes such as \cite{ICDCS20:Zeng} and \cite{INFOCOM21:Deng} start to concentrate on the performance improvement of FL, besides the properties of IR, IC, fairness, etc. 
    \item Most of studies use complicated techniques such as reinforcement learning, blockchain, Shapley value, which introduces large computational cost and communication overhead. The complicated and expensive computation is not suitable for mobile scenarios and resource-constraint participants.  
    \item Those studies which adopt a single-dimensional metric might not be appropriate for FL scenarios. Both resources provided by participants and model performance should be multi-dimensional in FL. 
    \item Most current mechanisms only targets at a single aspect of incentive scheme, and few studies involve node selection, payment allocation, and contribution evaluation simultaneously. However, these sub-problems are correlated, and should be considered together from a systemic point of view. 
\end{itemize}

\subsection{Future Studies of Incentive in FL}
In FL, the study of incentive mechanism is still in its infancy, and there exist many open issues. We just point out some directions for the future studies. 
\begin{enumerate}[(1)]
    \item The future incentive scheme should aim at improving the performance of FL with low costs by inspiring more participants to join in FL. One point is the final goal of incentive mechanism is definitely to improve the performance of FL. Otherwise, it is useless to design incentive schemes, even though they can inspire more participants. Another key point is that the proposed incentive scheme should be lightweight, since resource-constraint nodes hesitate to perform expensive computation of incentive method. 
    \item Researchers should put more emphasis on cross-silo FL. The decision behavior of large company/organization is distinct from that of end users and mobile devices, which further requires a totally new incentive method for cross-silo FL. Moreover, FL has the widespread applications in cross-silo setting, making incentive schemes more indispensable and critical. 
    \item Comprehensive incentive schemes are required for FL. This indicates that we should consider a multi-dimensional metric instead of a single-dimensional metric. In addition, future studies also appeal to multi-goals and multi-functionalities in one scheme. 
    \item Some cutting-edge technologies, like graph neural networks, generative adversarial networks, multi-agent reinforcement learning, etc., might find its potential application in the incentive design of new scenarios such as mobile edge computing and 5G/B5G. 
\end{enumerate}

\section{Conclusion}
In this survey, we provide a comprehensive introduction to the incentive mechanism design in FL scenarios. We formulate the incentive problem and present a taxonomy and some design goals of incentive mechanisms. The existing studies are summarized according to the main techniques they adopt, i.e., Shapley value, Stackelberg game, auction, contract theory, reinforcement learning, blockchain, etc. For each technique, we not only briefly present the brief introduction but also present the main problems in incentive design. After reviewing the current results, we compare them in several aspects and then point out four future directions, i.e., performance improvement of FL with low cost, incentive schemes in cross-silo setting, comprehensive incentive design, and the adoption of cutting-edge technologies in new scenarios.


\begin{thebibliography}{00}
\bibitem{Google17:McMahan} H. McMahan, E. Moore, D. Ramage, and B. Arcas, ``Federated Learning of Deep Networks using Model Averaging,'' \emph{arXiv preprint arXiv:1602.05629v2}, 2017.
\bibitem{ComSurvey21:Gu} R. Gu, C. Niu, F. Wu, G. Chen, C. Hu. C. Lu, and Z. Wu, ``From Server-based to Client-based Machine Learning: A Comprehensive Survey,'' \emph{ACM Computing Survey}, 2021.
\bibitem{TIST19:Yang} Q. Yang, Y. Liu, T. Chen, and Y. Tong, ``Federated Machine Learning: Concept and Applications,'' \emph{ACM Transactions on Intelligent Systems and Technology (TIST)}, vol. 10, no. 2, pp. 1201-1215, 2019.
\bibitem{Others18:Hard} A. Hard, K. Rao, R. Mathews, F. Beaufays, S. Augenstein, H. Eichner, C. Kiddon, and D. Ramage, ``Federated Learning for Mobile Keyboard Prediction,'' \emph{arXiv preprint arXiv:1811.03604}, 2018. 
\bibitem{Others19:Kairouz} P. Kairou, H. McMahan, B. Avent, A. Bellet, M. Bennis, A. Bhagoji, et al., ``Advances and Open Problems in Federated Learning,'' \emph{arXiv preprint arXiv:1912.04977}, 2019.
\bibitem{INFOCOM21:Tang} M. Tang and V. Wong,  ``An Incentive Mechanism for Cross-silo Federated Learning: A Public Goods Perspective,'' in Proc. of \emph{IEEE INFOCOM}, 2021.
\bibitem{INFOCOM19:Wang} Z. Wang, M. Song, Z. Zhang, Y. Song, Q. Wang, and H. Qi, ``Beyond Inferring Class Representatives: User-level Privacy Leakage From Fed- erated Learning,'' in Proc. of \emph{IEEE INFOCOM}, 2019.
\bibitem{ICDCS20:Zeng} R. Zeng, S. Zhang, J. Wang, and X. Chu, ``FMore: An Incentive Scheme of Multi-dimensional Auction for Federated Learning in MEC,'' in Proc. of \emph{IEEE ICDCS}, 2020.
\bibitem{FLWorkshop19:Cong} M. Cong, X. Weng, H. Yu, and Z.Qu, ``FML Incentive Mechanism Design: Concepts, Basic Settings, and Taxonomy,'' in Proc. of \emph{IEEE the 1st International Workshop on Federated Learning for User Privacy and Data Confidentiality}, 2019.
\bibitem{Others20:Cong} M. Cong, H. Yu, X. Weng, J. Qu, Y. Liu, and S. Yiu, ``A VCG-based Fair Incentive Mechanism for Federated Learning,'' \emph{arXiv preprint arXiv:2008.06680v1}, 2020.
\bibitem{AIES20:Yu} H. Yu, Z. Liu, Yang Liu, T. Chen, M Cong, X. Weng, D. Niyato, and Q. Yang, ``A Fairness-aware Incentive Scheme for Federated Learning,'' in Proc. of \emph{AAAI/ACM Conference on AI, Ethics, and Society (AIES)}, 2020.
\bibitem{INFOCOM21:Ding} N. Ding, Z. Fang, L. Duan, and J. Huang, ``Incentive Mechanism Design for Distributed Coded Machine Learning,'' in Proc. of \emph{IEEE INFOCOM}, 2021.
\bibitem{IoTJ20:Lim} W. Lim, Z. Xiong, C. Miao, D. Niyato, Q. Yang, C. Leung, and H. Poor, ``Hierarchical Incentive Mechanism Design for Federated Machine Learning in Mobile networks," \emph{IEEE Internet of Things Journal}, vol. 7, no. 10, pp. 9575-9588, 2020.
\bibitem{Others19:Kang} J. Kang, Z. Xiong, D. Niyato, H. Yu, Y. Liang, and D. Kim, ``Incentive Design for Efficient Federated Learning in Mobile Networks: A Contract Theory Approach,'' \emph{arXiv preprint arXiv:1905.07479}, 2019.
\bibitem{IoTJ:Kang} J. Kang, Z. Xiong, D. Niyato, S. Xie, and J. Zhang, ``Incentive Mechanism for Reliable Federated Learning: A Joint Optimization Approach to Combining Reputation and Contract Theory,'' \emph{IEEE Internet of Things Journal}, vol. 6, no. 6, pp.10700-10714, 2019. 
\bibitem{INL19:Sarikaya} Y. Sarikaya and O. Ercetin, ``Motivating Workers in Federated Learning: A Stackelberg Game Perspective'', \emph{IEEE Networking Letters}, 2020.
\bibitem{IoTJ20:Zhan} Y. Zhan, P. Li, Z. Qu, D. Zeng, and S. Guo, ``A Learning-based Incentive Mechanism for Federated Learning,'' \emph{IEEE Internet of Things Journal}, vol. 7, no. 7, pp. 6360-6368, 2020.
\bibitem{Others18:Feng} S. Feng, D. Niyato, P. Wang, et al., ``Joint Service Pricing and Cooperative Relay Communication for Federated Learning,'' \emph{arXiv preprint arXiv:1811.12082 }, 2018.
\bibitem{TMC20:Jiao} Y. Jiao, P. Wang, D. Niyato, B. Lin, and D. Kim, ``Toward an Automated Auction Framework for Wireless Federated Learning Services Market,'' \emph{IEEE Transactions on Mobile Computing}, 2020.
\bibitem{WCNC20:Le} T. Le, N. Tran, Y. Tun, Z. Han, and C. Hong, ``Auction Based Incentive Design for Efficient Federated Learning in Cellular Wireless Networks,'' in Proc. of \emph{IEEE Wireless Communications and Networking Conference (WCNC)}, 2020.  
\bibitem{WiOpt20:Ding} N. Ding, Z. Fang, and J. Huang, ``Incentive Mechanism Design for Federated Learning with Multi-Dimensional Private Information,'' in Proc. of \emph{International Symposium on Modeling and Optimization in Mobile, Ad Hoc, and Wireless Networks (WiOPT)}, 2020.
\bibitem{BigData19:Wang} G. Wang, C. Dang, and Z. Zhou, ``Measure Contribution of Participants in Federated Learning,'' in Proc. of\emph{IEEE International Conference on Big Data}, 2019.
\bibitem{IJCAI20:Ng} K. Ng, Z. Chen, Z. Liu, H. Yu, Y. Liu, and Q. Yang, ``A Multi-player Game for Studying Federated Learning Incentive Schemes,'' in Proc. of \emph{IEEE IJCAI}, 2020.
\bibitem{BigDate:Song} T. Song, Y. Tong and S. Wei, ``Profit Allocation for Federated Learning,'' in Proc. of \emph{IEEE International Conference on Big Data (BigData)}, 2019.
\bibitem{Others20:LiuYuan} Y. Liu, S. Sun, Z. Ai, S. Zhang, Z. Liu, and H. Yu, ``FedCoin: A Peer-to-Peer Payment System for Federated Learning,'' \emph{arXiv preprint arXiv:2002.11711v1}, 2020.
\bibitem{INFOCOM20:Wang} H. Wang, Z. Kaplan, D. Niu, and B. Li, ``Optimizing Federated Learning on Non-IID Data with Reinforcement Learning,'' in Proc. of \emph{IEEE INFOCOM}, 2020.
\bibitem{GLOBECOM20:Nishio} T. Nishio, R. Shinkuma, and N. Mandayam, ``Estimation of Individual Device Contributions for Incentivizing Federated Learning,'' in Proc. of \emph{IEEE Globecom}, 2020.
\bibitem{Others20:Liu} Y. Liu and J. Wei, ``Incentives for Federated Learning: A Hypothesis Elicitation Approach,'' \emph{arXiv preprint arXiv:2007.10596v1}, 2020.
\bibitem{GLOBECOM19:Shashi} S. Pandey, N. Tran, M. Bennis, and Y. Kyaw, ``Incentivize to Build: A Crowdsourcing Framework for Federated Learning,'', in Proc. of \emph{IEEE Globecom}, 2019.
\bibitem{Others19:Khan} L. Khan, S. Pandey, N. Tran, W. Saad, Z. Han, M. Nguyen, and C. Hong, ``Federated Learning for Edge Networks: Resource Optimization and Incentive Mechanism,'' \emph{arXiv preprint arXiv:1911.05642}, 2019.
\bibitem{Others20:Lim} W. Lim, J. Huang, Z. Xiong, et al., ``Towards Federated Learning in UAV-Enabled Internet of Vehicles: A Multi-Dimensional Contract-Matching Approach,'' \emph{arXiv preprint arxiv:2004.03877}, 2020.
\bibitem{INFOCOM2015:Zhao} J. Zhao, X. Chu, H. Liu, Y. Leung, and Z. Li, ``Online Procurement Auctions for Resource Pooling in Client-Assisted Cloud Storage Systems,'' in Proc. of \emph{IEEE INFOCOM}, 2015.
\bibitem{AISTATS19:Jia} R. Jia, D. Dao, B. Wang, et al., ``Towards Efficient Data Valuation Based on Shapley Value,'' in Proc. of \emph{IEEE 22nd International Conference on Artificial Intelligence and Statistics (AISTATS)}, 2019. 
\bibitem{Others19:Ghorbani} A. Ghorbani and J. Zou, ``Data Shapley: Equitable Valuation of Data for Machine Learning,'' \emph{arXiv preprint arXiv:1904.02868v2}, 2019.
\bibitem{INFOCOM18:Yu} H. Yu, G. Iosifidis, B. Shou, and J. Huang, ``Market Your Venue with Mobile Applications: Collaboration of Online and Offline Business,'' in Proc. of \emph{IEEE INFOCOM}, 2018.
\bibitem{INFOCOM19:Yu} H. Yu, E. Wei, and R. Berry, ``A Business Model Analysis of Mobile Data Rewards,'' in Proc. of \emph{IEEE INFOCOM}, 2019.
\bibitem{ComSurveys11:Parsons} S. Parsons, J. Aguilar, and M. Klein, ``Auctions and Bidding,'' \emph{ACM Computing Surveys}, vol. 43, no. 2, pp. 1-66, 2011.
\bibitem{ICML19:Duetting} P. Duetting, Z. Feng, H. Narasimhan, D. Parkes, and S. Ravindranath, ``Optimal Auctions Through Deep Learning,'' in Proc. of \emph{IEEE ICML}, 2019.
\bibitem{SPM17:Arulkumaran} K. Arulkumaran, M. Deisenroth, M. Brundage, and A. Bharath, ``Deep Reinforcement Learning: A Brief Survey,'' \emph{IEEE Signal Processing Magazine}, vol. 34, no. 6, pp. 26-38, 2017.
\bibitem{Others17:Hessel} M. Hessel, J. Modayil, H. Hasselt, T. Schaul, et al., ``Rainbow: Combing Improvements in Deep Reinforcement Learning,'' \emph{arXiv preprint arXiv:1710.02298v1}, 2017.
\bibitem{CST19:Yang} R. Yang, F. Yu, P. Si, Z. Yang, and Y. Zhang, ``Integrated Blockchain and Edge Computing Systems: A Survey, Some Research Issues and Challenges,'' \emph{IEEE Communications Surveys \& Tutorials}, vol. 21, no. 2, pp. 1508-1532, 2019.
\bibitem{BigData19:Toyoda} K. Toyoda and A. Zhang, ``Mechanism Design for An Incentive-aware Blockchain-enabled Federated Learning Platform,'' in Proc. of \emph{IEEE International Conference on Big Data (Big Data)}, 2019.
\bibitem{BigCom2019:Bao} X. Bao, C. Su, Y. Xiong, W. Huang, and Y. Hu, ``FLChain: A Blockchain for Auditable Federated Learning with Trust and Incentive,'' in Proc. of \emph{IEEE BigCom}, 2019.
\bibitem{ICDMW18:Lu} Y. Lu, Q. Tang, and G. Wang, ``On Enabling Machine Learning Tasks atop Public Blockchains: A Crowdsourcing Approach,'' in Proc. of \emph{IEEE International Conference on Data Mining Workshops (ICDM)}, 2018.
\bibitem{TDSC19:Weng} J. Weng, J. Weng, J. Zhang, M. Li, Y. Zhang, and W. Luo, ``Deepchain: Auditable and Privacy-preserving Deep Learning with Blockchain-based Incentive,'' \emph{IEEE Transactions on Dependable and Secure Computing}, 2019.
\bibitem{Others18:Nishio} T. Nishio and R. Yonetani, ``Client Selection for Federated Learning with Heterogeneous Resources in Mobile Edge,'' \emph{arXiv preprint arXiv:1804.08333v2}, 2018.
\bibitem{IIS:Yu} H. Yu, Z. Liu, Y. Liu, T. Chen, M. Cong, X. Weng, D. Niyato, and Q. Yang, ``A Sustainable Incentive Scheme for Federated Learning,'' \emph{IEEE Intelligent Systems}, 2019.
\bibitem{Others20:Yan} B.Yan, Y. Zhou, J. Wang, L. Liu, Y. Zhang, and X. Nie, ``A Real-time Contribution Measurement Method for Participants in Federated Learning,'' \emph{arXiv preprint arXiv:2009.03510}, 2020.
\bibitem{Others20:Cai} H. Cai, D. Rueckert, and J. Palmbach, ``2CP-Decentralized Protocols to Transparently Evaluate Contributivity in Blockchain Federated Learning Environments,'' \emph{arXiv preprint arXiv:2011.07516v1}, 2020. 
\bibitem{Others20:Chen} M. Chen, Y. Liu, W. Shen, Y. Shen, P. Tang, and Q. Yang, ``Mechanism Design for Multi-Party Machine Learning,'' \emph{arXiv preprint arXiv:2001.08996}, 2020.
\bibitem{GLOBECOM20:Hu} R. Hu, Y.Gong, ``Trading Data For Learning: Incentive Mechanism for On-Device Federated Learning,'', in Proc. of \emph{IEEE Globecom}, 2020.
\bibitem{INFOCOM21:Weng} J. Weng, J. Weng, H. Huang, C. Cai, and C. Wang, ``FedServing: A Federated Prediction Serving Framework Based on Incentive Mechanism,'' in Proc. of \emph{IEEE INFOCOM}, 2021.
\bibitem{INFOCOM21:Deng} Y. Deng, F. Lyu, J. Ren, Y. Chen, P. Yang, Y. Zhou, and Y. Zhang, ``FAIR: Quality-Aware Federated Learning with Precise User Incentive and Model Aggregation,'' in Proc. of \emph{IEEE INFOCOM}, 2021.
\bibitem{NIPS17:Lundberg} S. Lundberg and S Lee, ``A Unified Approach to Interpreting Model Predictions,'' in Proc. of \emph{NIPS}, 2017. 
\bibitem{Networks20:Zhan} Y. Zhan, P. Li, S. Guo, and Z. Qu, ``Incentive Mechanism Design for Federated Learning: Challenges and Opportunites,'' \emph{IEEE Networks}, 2021. 
\bibitem{TETC21:Zhan} Y. Zhan, J. Zhang, Z. Hong, L. Wu, P. Li, and S. Guo, ``A Survey of Incentive Mechanism Design for Federated Learning,'' \emph{IEEE Transactions on Emerging Topics in Computing}, 2021.
\end{thebibliography}
\end{document}